\documentclass{article}

\usepackage{arxiv}

\usepackage[utf8]{inputenc} 
\usepackage[T1]{fontenc}    
\usepackage[linktocpage]
            {hyperref}       
\usepackage{url}            
\usepackage{booktabs}       
\usepackage{comment}		
\usepackage{amsfonts}       
\usepackage{nicefrac}       
\usepackage{microtype}      
\usepackage{lipsum}			
\usepackage{graphicx}
\usepackage{natbib}
\usepackage{doi}
\usepackage{mathtools}
\usepackage{amsmath} 
\usepackage{verbatim}
\usepackage{color}
\usepackage{paralist}       
\usepackage{amsthm}         
\usepackage{wrapfig}        
\usepackage{bm}
\usepackage{tocloft}     
\usepackage{etoc}
\etocsettocstyle{}{}

\DeclareMathSymbol{*}{\mathbin}{symbols}{"01}

\usepackage{tikz}
\usetikzlibrary{shapes.geometric, arrows}

\tikzstyle{startstop} = [rectangle, rounded corners, minimum width=3cm, minimum height=1cm,text centered, draw=black, fill=red!30]
\tikzstyle{process} = [rectangle, minimum width=3cm, minimum height=1cm, text centered, draw=black, fill=orange!30]
\tikzstyle{decision} = [diamond, minimum width=3cm, minimum height=1cm, text centered, draw=black, fill=green!30]
\tikzstyle{arrow} = [thick,->,>=stealth]

\usetikzlibrary{trees,shapes.geometric,arrows.meta}

\tikzset{
decision/.style = {rectangle, rounded corners, draw, align=center, text width=1.7cm, fill=gray!10},
decisionBIG/.style = {rectangle, rounded corners, draw, align=center, text width=2.4cm, fill=gray!40},
decisionMittel/.style = {rectangle, rounded corners, draw, align=center, text width=1.8cm, fill=gray!25},
leaf/.style = {rectangle, draw=none, align=center, text width=1.3cm, fill=gray!10},
    root/.style = {ellipse, draw, align=center, text width=2.7cm},
    edge from parent/.style={draw,-{stealth},thick},
}

\newtheorem{Thm}{Theorem}[section]

\theoremstyle{definition} 

\newtheorem{Def}[Thm]{Definition}

\newtheorem{Rem}[Thm]{Remark}

\title{Absolute Evaluation Measures for Machine Learning:\\ A Survey}

\author{\medskip 
Silvia Beddar-Wiesing\textsuperscript{1,\dag},  Alice Moallemy-Oureh\textsuperscript{1,\dag}, Marie Kempkes\textsuperscript{1,\dag}, Josephine M.~Thomas\textsuperscript{2}\\
	$^1$ Intelligent Embedded Systems, University of Kassel\\\medskip
	\texttt{marie.kempkes@outlook.de}, \texttt{\{amoallemy,  s.beddarwiesing\}@uni-kassel.de} \\
    $^2$ Institute of Data Science, University of Greifswald\\
    \texttt{thomasj@uni-greifswald.de}
}

\renewcommand{\shorttitle}{Survey on Absolute Evaluation Measures}

\begin{document}
\maketitle

\footnotetext[1]{These authors contributed equally.}
\begin{abstract}
Machine Learning is a diverse field applied across various domains such as computer science, social sciences, medicine, chemistry, and finance. This diversity results in varied evaluation approaches, making it difficult to compare models effectively. Absolute evaluation measures offer a practical solution by assessing a model's performance on a fixed scale, independent of reference models and data ranges, enabling explicit comparisons. However, many commonly used measures are not universally applicable, leading to a lack of comprehensive guidance on their appropriate use. This survey addresses this gap by providing an overview of absolute evaluation metrics in ML, organized by the type of learning problem. While classification metrics have been extensively studied, this work also covers clustering, regression, and ranking metrics. By grouping these measures according to the specific ML challenges they address, this survey aims to equip practitioners with the tools necessary to select appropriate metrics for their models. The provided overview thus improves individual model evaluation and facilitates meaningful comparisons across different models and applications.

\end{abstract}

\section*{Introduction}
Evaluating Machine Learning (ML) models presents numerous challenges due to the diversity of methods used to tackle different learning problems, such as classification, regression, clustering, and ranking. Each problem type necessitates distinct learning algorithms, error functions, and evaluation procedures. Additionally, the nature of the dataset may demand tailored evaluation methods. 

First, clarifying the distinction between "performance measures" and "evaluation measures" is crucial. Oftentimes, both terms are used interchangeably. However, performance measures are particularly discussed in fields like finance \cite{caporin2014survey, stede2006strategy, chenhall2007multiple, tangen2003overview} or Reinforcement Learning \cite{fang2017learning, datar2001balancing, szepesvari2010algorithms}, assessing the success of an agent based on specific criteria or rewards. Evaluation measures such as accuracy, on the other hand, are more general functions used in ML to estimate the overall model performance. This survey focuses on the latter, aiming to determine the most appropriate evaluation functions for a given context. Choosing the correct measure is foundational for comparing models and drawing meaningful conclusions.

However, using varied and sometimes inappropriate evaluation measures within the same domain can complicate comparisons in ML. As a result, determining the most suitable measure for a specific context requires careful consideration. Given these complexities, this survey narrows its focus to a fundamental question: 

\begin{center}
    \textit{How well does a single ML model perform for a specific data set and learning problem?}
\end{center}

This question can be effectively addressed using \textbf{absolute measures}. They are particularly useful in Machine Learning as their output is confined to a fixed interval. This allows for consistent evaluation and meaningful comparison of models across different datasets and conditions.

The literature on evaluation measures in ML is extensive, including reviews of not necessarily absolute evaluation measures in ML \cite{rainio2024evaluation, naidu2023review} and topics such as general performance evaluation \cite{japkowicz2015performance}, the creation of new measures \cite{huang2007constructing}, and their application in specific learning problems or imbalanced datasets. Various surveys have explored these areas, comparing evaluation measures for classification \cite{sokolova2009systematic, hossin2015review, vujovic2021classification, ferri2009experimental, flach2019performance, garcia2009study}, unsupervised learning \cite{palacio2019evaluation}, imbalanced data \cite{fatourechi2008comparison, kulatilleke2022empirical, yu2021survey, krawczyk2016learning}, and more. In addition, the explainability of evaluation measures  \cite{survey_gnn_explainability_evaluation, wang2022unified} and measures for ML explanations \cite{zhou2021evaluating} have been investigated.

Furthermore, many surveys have been published about suitable evaluation measures and their interpretation possibilities throughout investigating different applications. For example, for medical image segmentation, \cite{muller2022towards} provides an overview and interpretation guide on standard evaluation measures in binary and multi-class problems. Moreover, in different areas of Natural Language Processing, as for text comprehension \cite{survey_reading_comprehension}, text mining \cite{dalianis2018evaluation}, linguistics \cite{steen2022find},  
style transfer \cite{briakou2021evaluating},
and natural language generation \cite{sai2022survey},
corresponding overviews and discussions of evaluation measures are provided. Furthermore, surveys are available on evaluation measures for Neural Networks \cite{twomey1995performance}, 
recommender systems \cite{gunawardana2009survey}, web prefetching systems \cite{survey_web_prefetching_performance}, 
single object tracking \cite{soleimanitaleb2022single}, target tracking \cite{song2022performance}, and for classification and regression tasks in engineering and sciences \cite{naser2023error}, etc.

However, many existing measures are not absolute, limiting their applicability to compare models for different data and application settings due to a missing common baseline for comparison.  
Thus, this survey provides an overview of absolute measures for classification, clustering, and ranking tasks that enable comparing models independent of the dataset, architecture, and other factors. 
This survey discusses the different metrics in detail to provide an intuition for the reader to select appropriate measures for specific ML tasks. In addition, metrics for domain-specific tasks are discussed briefly. The compendium of the proposed discussions is illustrated in decision trees, which serve as concise and easily accessible guidelines for proper choices of evaluation measures, making evaluations in ML more consistent and comparable.

This survey is organized as follows\footnote{This survey includes literature published up to January 2025. Research that emerged after this point is not considered within the scope of this work.}. After introducing the different ML problems, evaluation metrics, and several of their properties in Sec.~\ref{section_preliminaries}, the survey consists of Sec.~\ref{section_classification} - \ref{section_ranking}, one for each learning problem and a last one for domain-specific measures in Sec.~\ref{section_data-specific}. Each section ends with a discussion of the described evaluation measures and the illustration of the task-specific decision tree. Finally, we conclude in Sec.~\ref{section_discussion} with an extensive discussion on the usability of the listed measures.
\tableofcontents
\newpage
\section{Preliminaries}\label{section_preliminaries}
Depending on the learning problem, the evaluation metrics for a particular model can be chosen. These problems include classification, clustering, regression, and ranking tasks. 
In the following, these tasks are briefly introduced. Afterward, in Subsec.~\ref{section_performance_metrics}, an introduction to absolute evaluation measures is given. In this context, the definition of the confusion matrix is also revised, as it is the basis for many evaluation measures, especially for classification problems.

\subsection{Learning Problems}\label{section_learning_problems}

This section briefly introduces learning problems that different ML methods can solve. 
Depending on the given information, i.e., the dataset, different types of learning problems can be studied. 
These problems are categorized as supervised, semi-supervised, or unsupervised learning problems. The definitions of the following learning problems are based on \cite{book_bishop_2006}, \cite{book_goodfellow_2016}, \cite{book_hastie_2009} and \cite{book_vapnik_2000}. 

In general, a learning problem is said to be
\begin{compactenum}
\item[•] \textbf{supervised} if the data set to be explored provides fully-labeled data,
\item[•] \textbf{unsupervised} if the data set consists of only unlabeled data and
\item[•] \textbf{semi-supervised} if the data set provides partially labeled data.
\end{compactenum}

The classic learning problems that can be divided into these three areas include the following tasks.

\paragraph{Classification} requires the knowledge of different classes into which the data set can be divided. Given a similarity measure $\phi$, data objects within the same class are assumed to be more similar w.r.t.~$\phi$ than objects in different classes. The classification learning problem consists of assigning a class to objects not yet classified. Note that by requiring knowledge of class membership only from the training data, this learning problem can be both supervised and semi-supervised.

\paragraph{Clustering} can be interpreted as a form of classification from an unsupervised point of view.
Given a similarity measure $\phi$, clustering defines the task of assigning the data points to different classes such that data points within the same class are more similar w.r.t.~$\phi$ than data of different classes. If the class labels are given, the clustering is called supervised, and unsupervised otherwise. For more information on clustering, see \cite[§12, §14]{book_hastie_2009}.

\paragraph{Ranking} involves ordering the input data optimally according to specific criteria \cite{jour_cohan_1999, jour_agarval_2005}. The ordering criteria must be known for learning, so Ranking is a semi-supervised learning problem.

\paragraph{Note.} Another common task is regression, which involves learning a function that maps input data to continuous outputs \cite[§3]{book_bishop_2006}, \cite[§5]{book_goodfellow_2016}, \cite[§2, §3, §6, §9, §11]{book_hastie_2009}. 
While regression is typically evaluated using residual-based loss functions (e.g., RMSE, MSE, MAE), they are not absolute. However, when interpreted as a ranking problem, regression tasks can be evaluated using correlation metrics \cite{japkowicz2011evaluating}. 
Additionally, since correlation metrics are relatively uncommon evaluation methods, we neither cover the regression task nor correlation metrics further in this survey.

 \subsection{Absolute Evaluation Measure}\label{section_performance_metrics}

For a clear understanding of this survey, we first introduce the concept of evaluation measures used in this survey and describe their properties. 
While covering all possible measures is beyond our scope, we focus on absolute metrics that evaluate a single model's effectiveness in addressing a specific problem. Additionally, we discuss other properties of evaluation measures that can influence the results.

It is important to note that the same evaluation measures may be defined differently across various sources. For instance, \cite[§3.1.1 and Table 1]{survey_web_prefetching_performance} shows that terms like \textbf{precision}, \textbf{recall}, and \textbf{accuracy} may have varying definitions. We use the most common or intuitive term for each metric presented to ensure clarity and readability.

\begin{Def}[Absolute Evaluation Measure]
Let $I = [a, b]$ be an interval of evaluation values of a model, i.e., a model performs better when $i\in I$ is close to $a$ and worse the closer $i$ is to $b$, or vice versa. For a learning problem $\mathcal{L}$ on a dataset $\mathcal{D}$ and an ML model $\mathcal{M}$ an absolute evaluation measure is a function $\phi$ with  $(\mathcal{L}, \mathcal{D}, \mathcal{M}) \longmapsto i \in I$. 
\end{Def}

\subsection{Confusion Matrix}\label{section_confusion_matrix}

Intuitive evaluation measures, often used in classification problems, are typically based on the confusion matrix \cite{japkowicz2011evaluating}. Some metrics incorporate additional factors, such as prior class distribution, classifier uncertainties, or weighted errors. To provide a clearer understanding of the strengths and weaknesses of the measures discussed in Sec.\ref{section_metrics_classification}, we present an overview of the confusion matrix and key properties of these evaluation measures \cite{japkowicz2011evaluating}.

\begin{Def}[Confusion Matrix]
Let $f$ be a classifier and $\{1, \ldots, C\}$ the set of classes. Then the $i,j$-th entry of the confusion matrix $\mathcal{C}(f)\in \mathbb{N}^{C \times C}$ is defined as
\begin{equation}\label{eq_confusion}
    c_{ij}(f) := \sum_{x \in \mathcal{D}} [(y(x) = i) \wedge (f(x) = j )],
\end{equation}
where $i, j \in \{1, \ldots, C\}$ are class labels, $x \in \mathcal{D}$ are test examples from a test set $\mathcal{D}$ with $y(x)$ is its true label and $f(x)$ the predicted label. So the matrix entry $c_{i,j}$ corresponds to the number of examples belonging to class $i$ but getting classified into class $j$.
\end{Def}

The confusion matrix helps determine whether a dataset is balanced or imbalanced toward one of the classes (positive or negative). 
With this insight, one can choose the most appropriate metric for model evaluation. A notable special case occurs when $C = 2$, corresponding to the \textit{binary classification problem}, where the confusion matrix entries have specific names:
\begin{align}
    &c_{1,1} : \text{true positive (TP)} \quad c_{1,2} : \text{false negative (FN)} \nonumber\\ 
    &c_{2,1} : \text{false positive (FP)} \quad c_{2,2} : \text{true negative (TN)}\label{fig:ConfMatrix}
\end{align}

A true positive occurs when the input belongs to the positive class, and the ML model correctly classifies it as positive. Similarly, a true negative occurs when a negative class input is correctly classified as negative. If the model incorrectly classifies a negative input as positive, it is called false positive. Conversely, a false negative occurs when a positive input is wrongly classified as negative \cite{zeng2020confusion}. These incorrect classifications are also known as type-1 and type-2 errors, respectively. In this context, an evaluation measure is called \textbf{type-1 error appropriate} if the model prioritizes minimizing the type-1 errors over the type-2 errors, and \textbf{type-2 error appropriate} if the model prioritizes minimizing the type-2 errors over the type-1 errors.

\subsection{Data Balance and Chance Correction}

In many applications learning a task on a data set faces the problem of imbalances. \textbf{Balanced data} is almost evenly distributed across the classes which forces a model to perform well on every class in many learning approaches. However, \textbf{imbalanced data} includes classes of different sizes, skewing the focus of an ML model during training. In the extreme case, one class is highly underrepresented and the performance of a model is already considered as sufficient if it classifies the substantially overrepresented classes well while neglecting the underrepresented classes. Various evaluation metrics for imbalanced data have been developed to prevent this and expose the fact that the model only works well for specific classes and not all classes.

Furthermore, verifying whether the model works better than uniformly distributed random predictions can be helpful, i.e., better than performance by chance. For that purpose, an evaluation measure must be \textbf{chance-corrected}, which means that the performance by chance is explicitly deducted. Many measures for balanced single-class data are inherently chance-corrected, whereas measures for imbalanced multi-class data are explicitly modeled as chance-corrected. As a result, the measures are often more complex to interpret and apply, but they allow interpretations concerning random predictions.
\label{section_metrics_classification}

\section{Classification Measures}\label{section_classification}

Classification tasks offer a diverse set of evaluation measures that can be categorized based on the characteristics of the data, such as whether it is balanced or imbalanced and whether the task involves single-class or multi-class classification. Some measures also include a chance correction to adjust for biases inherent in the labeling process or the model itself.

This section begins with an overview of classification measures applicable to balanced single-class data, discussing their advantages and disadvantages. Subsequently, we explore measures suited for balanced multi-class data. Then, measures for imbalanced single- and multi-class data are described, with some allowing for chance correction, eliminating the probability that the model classifies randomly. Finally, we provide an overview of the absolute classification measures in a concept tree.

\subsection{Measures on Balanced Data with Single Class}\label{sec_measures_balanced_single_class}

Binary classification on balanced data allows for various measures based on the confusion matrix \cite{japkowicz2011evaluating}. Here, class distributions are assumed to be uniform, meaning labels are approximately equally represented in a single-class context.

One of the most common measures for balanced single-class data is the \textbf{accuracy} ($acc$), which calculates the proportion of correctly classified samples to all samples \cite{book_bishop_2006}. It is defined as the ratio of correct classifications ($TP + TN$) to the total number of data points:
\begin{align}\label{accuracy}
acc = \frac{TP + TN}{TP + TN + FP + FN}.
\end{align}
In the confusion matrix, cf.~Eq.~\eqref{fig:ConfMatrix}, this corresponds to dividing the sum of the diagonal elements by the sum of all entries. The complement to accuracy is the \textbf{error rate}, also known as Brier score in the binary case~\cite{chicco2021thematthews}.

It is essential to keep in mind that different measures emphasize various types of errors. Consequently, the choice and interpretation of these measures are crucial in evaluation. The following measures are categorized based on their suitability for type-1 and type-2 errors and adjustable ones.

A common type-1 error-appropriate is \textbf{precision}, also known as \textbf{positive predictive value (PPV)}. Precision is calculated as the ratio of true positives to the total number of positive predictions:
\begin{align}\label{precision}
p = \frac{TP}{TP + FP}.
\end{align}
In Eq.~\eqref{fig:ConfMatrix}, this corresponds to dividing the upper left entry (correctly classified data points of the positive class) by the sum of the first row (all points classified as class positive) \cite{lantz2013machine}. Precision reflects the probability that a positive prediction by the model is correct. In the literature, in some cases, it is preferred to obtain a small value $1-p$, which is also known as \textbf{false discovery rate} \cite{benjamini1995controlling}.

Another type-1 error appropriate measure is the \textbf{true negative rate (TNR)}, also called \textbf{specificity} \cite{book_bishop_2006}. TNR reflects the proportion of actual negatives correctly identified by the model. It is calculated as:
\begin{align}\label{TNR}
TNR = \frac{TN}{TN + FP}.
\end{align}
In the confusion matrix shown in Eq.~\eqref{fig:ConfMatrix}, this is represented by dividing the lower right element (correctly classified as negative class) by the total of the second column (all data points of the negative class).

In contrast, type-1 errors are less critical in measures appropriate for type-2 errors (false negative classification). For example, misclassifying a sick person as healthy (False Negative) can be more severe than misclassifying a healthy person as sick. The former can have serious consequences, while the latter generally leads to additional, though unnecessary, tests.

A common measure in this context is \textbf{recall}, also known as \textbf{sensitivity}, \textbf{hit rate}, or \textbf{accuracy on the positive class}. Recall is equivalent to the true positive rate (TPR), as defined in Eq.~\eqref{Eq_Typ1_Typ2_error} \cite{powers2020evaluation}.

The \textbf{false positive rate ($FPR$)} is calculated from the confusion matrix in Eq.~\ref{fig:ConfMatrix} by dividing the upper right entry (false positives) by the total of the second column (data points of the negative class). It represents the probability that the model classifies a sample as positive when it is actually negative. A good $FPR$ is close to zero. While rarely used as a standalone metric, the $FPR$ is commonly paired with the \textbf{true positive rate} $TPR$ to generate the \textbf{receiver operating characteristic (ROC) curve}, discussed in Sec.~\ref{sec_measures_balanced_single_class}. 

\begin{equation}\label{Eq_Typ1_Typ2_error}
\begin{split}
    FPR = \frac{FP}{FP + TN}, \quad 
    TPR = \frac{TP}{TP+FN}, \quad
    FNR = \frac{FN}{FN + TP}.
\end{split}
\end{equation}

Similarly, the \textbf{false negative rate} $FNR = 1- TPR$ is obtained by dividing the lower left entry (false negatives) by the total of the first column (data points of positive class).

Another measure that focuses on minimizing type-2 error is the \textbf{negative predictive value (NPV)} \cite{book_bishop_2006}, which quantifies the probability that a sample classified as negative is actually negative:
\begin{align}\label{NPV}
NPV = \frac{TN}{TN + FN}.
\end{align}
NPV can be seen as a conditional probability given that the model predicted the negative class and can be intuitively understood as precision for the negative class.

In the context of balancing type-1 and type-2 errors, the \textbf{$\text{F}_\beta$-score} \cite{book_bishop_2006} offers a way to weigh both errors based on a parameter $\beta \in \mathbb{R}$. It is defined as:
\begin{equation}\label{f_beta}
    F_\beta = \frac{(1+\beta^2) \cdot p \cdot r}{\beta^2 \cdot p + r},
\end{equation}
where $p$ is the precision and $r$ is the recall. The F-score allows for adjusting the importance given to each error type. A high F-score indicates high precision and recall values, whereas a low F-score suggests that at least one of both is low, implying more type-1 or type-2 errors.

The F-score is useful when both types of errors need to be minimized, such as in loan approval, where false negatives (rejecting a creditworthy customer) and false positives (approving a risky customer) have significant implications. Typical $\beta$ values are:
\begin{enumerate}
\item $\text{F}_{0.5}$-score: Emphasizes precision over recall.
\item $\text{F}_{1}$-score: Balances precision and recall.
\item $\text{F}_{2}$-score: Emphasizes recall over precision.
\end{enumerate}

The \textbf{$\text{F}_{1}$-score}, the harmonic mean of precision and recall, is the most commonly used F-score \cite{torgo2009precision}. It is also known as the \textbf{sørensen–dice index} or \textbf{dice similarity coefficient} \cite{muller2022towards}.

\begin{wrapfigure}[30]{r}{0.3\textwidth}
\vspace{-1cm}
	\includegraphics[width=0.9\linewidth]{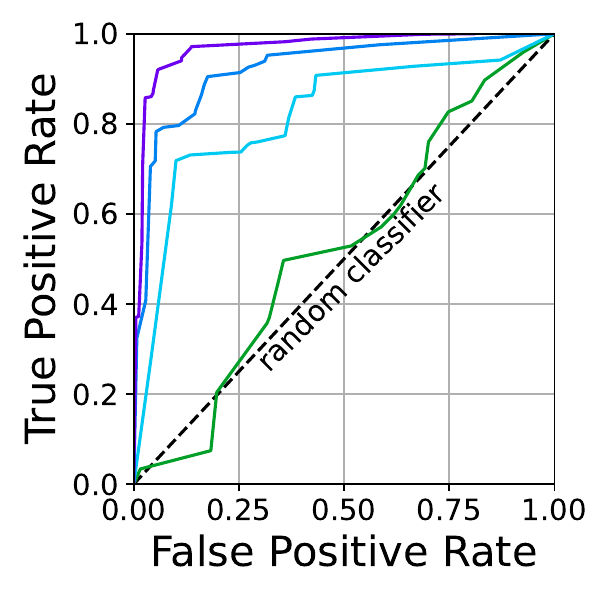}
	\caption{ROC curves for different classifiers. A perfect classifier has an FP rate of 0 and a TP rate of 1.}
	\label{fig:ROC_example}
	\vspace{0.4cm}
    \includegraphics[width=0.9\linewidth]{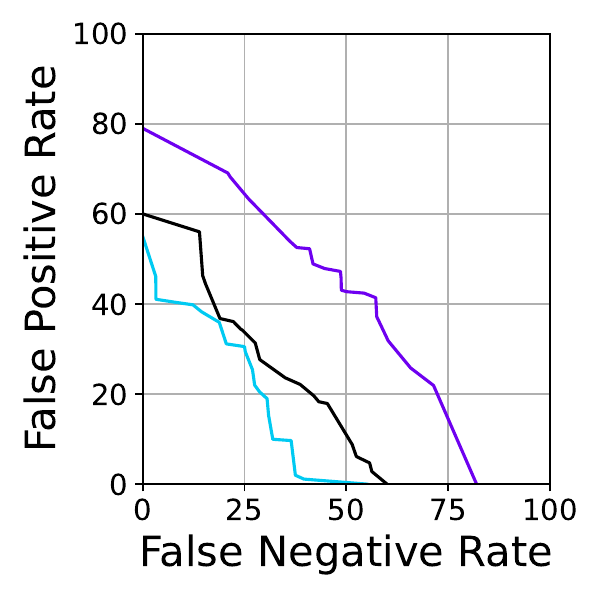}
	\caption{DET curve example \cite{Martin1997}. The red curve represents the best-performing model, while the blue and black curves show poorer classification ability.}
    \label{fig:DET_example}
\end{wrapfigure}

Graphical measures provide an intuitive way to evaluate and compare model performance by visualizing trade-offs between classification metrics across different thresholds.
Among these, one of the most commonly used graphical measures is the \textbf{area under the receiver operating characteristic curve (AUC-ROC)} \cite{book_bishop_2006}, which evaluates a model's performance across various classification thresholds. The ROC curve plots the true positive rate (TPR) against the false positive rate (FPR) by varying the decision threshold, as illustrated in Fig.~\ref{fig:ROC_example}. The AUC-ROC quantifies the overall performance, with a value close to 1 indicating a perfect classifier and 0.5 indicating a random classifier.

The \textbf{detection error trade-off (DET) curve} \cite{Martin1997} is another two-dimensional evaluation measure for classification models. It plots the false negative rate (FNR) against the false positive rate (FPR), as shown in Fig.~\ref{fig:DET_example}. Like the AUC-ROC, the DET curve helps evaluate model performance. 
It can be used as the AUC-ROC.

\paragraph{Discussion.}
Given the variety of measures used for single-class classification problems on balanced data, it is essential to consider which measure might be most suitable for specific contexts and whether any measure stands out as particularly effective.

The first consideration is whether type-1 or type-2 errors are of more significant concern. If both are equally important, measures such as accuracy, the $\text{F}_{\beta}$ score, or graphical tools might be appropriate. 
Accuracy provides a quick overview, but it can be misleading, especially with imbalanced datasets, as it tends to overestimate model performance \cite{ranawana2006optimized}. While it can indicate model adequacy in balanced scenarios, it should not be relied upon exclusively. Instead, alternative measures like balanced accuracy, as discussed in Sec.~\ref{section_classification_imbalanced_single_class}, are recommended when dealing with imbalanced data. Accuracy should thus be used cautiously, primarily in cases where data balance is assured.

For these reasons, the $\text{F}_{\beta}$-score is often preferred as it allows for adjusting the balance between model performances across classes. If no specific preference is given for type-1 or type-2 errors, or if both errors are equally important, the $\text{F}_{1}$-score is recommended due to its common use in such scenarios.

Graphical evaluation measures are often easier to interpret and more informative than non-graphical measures. For instance, the AUC-ROC, derived from a two-dimensional curve, provides a broader view of model performance compared to single-threshold measures like accuracy, especially in imbalanced datasets. However, the AUC-ROC can be overly optimistic for datasets with severe class imbalances \cite{gray2011further}. Therefore, Martin et al.\,\cite{Martin1997} suggest that the DET curve is more effective for comparison, as it frequently yields nearly linear results.

When evaluating type-1 error-appropriate measures, the choice between precision and true negative rate (TNR) depends on the specific context. Precision reflects the model's ability to avoid false positives (FPs) by measuring the proportion of true positives among all positive predictions. TNR, on the other hand, measures how well the model identifies true negatives (TN) and hence focuses on the correct classification of negative cases.

For type-2 error-appropriate measures, the decision between recall and negative predictive value (NPV) is similarly context-dependent. Recall aims to identify false negatives (FNs) by measuring the proportion of true positives among all actual positives. NPV, however, assesses how well the model identifies true negatives among all negative predictions, focusing on the correctness of negative classifications.

Neither precision, TNR, recall, nor NPV should be used in isolation. Instead, a combination of measures is often more informative. Relying solely on one measure may not provide a comprehensive view of model performance, particularly in cases where the data characteristics influence the relevance of different measures.

\subsection{Measures on Balanced Data with Multiple Classes}\label{sec_measures_balanced_multi_class}

Multi-class scenarios comprise data sets that are divided into more than two classes. Measures for classifying multi-class data cannot utilize the binary classification measures directly, but they can be extended directly using the multi-class confusion matrix from eq.~\eqref{eq_confusion} as listed in \cite{sokolova2009systematic}.

To extend accuracy for multi-class problems, we use the \textbf{average accuracy}, which calculates the mean accuracy across all classes \cite{sokolova2009systematic}. Formally, it is defined as:
\begin{align}\label{average_accuracy}
    acc_{\text{avg}} = \frac{1}{C}\sum_{i=1}^{C} \frac{TP_i + TN_i}{TP_i + FN_i + FP_i + TN_i},
\end{align}
where \( TP_i \), \( TN_i \), \( FP_i \), and \( FN_i \) represent the true / false positives and true / false negatives for each class \( i \in [C] \).

The \textbf{micro F$_\beta$-score} is a commonly used evaluation measure for multi-label classification where each class is equally important \cite{sokolova2009systematic}. A higher micro F$_\beta$-score, ranging from 0 to 1, indicates better performance. It is computed using \textbf{micro precision} \( P_\text{micro} \) and \textbf{micro recall} \( R_\text{micro} \) defined as follows:
\begin{align}\label{micro_precision_micro_recall}
    P_\text{micro} = \frac{\displaystyle\sum_{i=1}^C TP_i}{\displaystyle\sum_{i=1}^C (TP_i + FP_i)}, \quad
    R_\text{micro} = \frac{\displaystyle\sum_{i=1}^C TP_i}{\displaystyle\sum_{i=1}^C (TP_i + FN_i)}.
\end{align}
The \textbf{micro F$_\beta$-score} is then given as:
\begin{equation}\label{micro_f_beta}
    F_{\beta,\text{micro}} = \frac{(1+\beta^2) P_\text{micro} R_\text{micro}}{\beta^2 P_\text{micro} + R_\text{micro}}.
\end{equation}

\begin{Rem}\label{rem_graphical_extension}
Except for accuracy, the extended measures can be classified into those focusing on type-1 or type-2 errors. Precision remains a type-1 error-appropriate measure, while recall is a type-2 error-appropriate measure. 
Further, the ROC curve from the single-class classification problem, introduced in section~\ref{sec_measures_balanced_single_class}, can be extended to the multiclass case by using the multiclass versions of the precision and recall or by using the one-versus-all and one-versus-one approaches for the binary ROC curve, as illustrated in \href{https://scikit-learn.org/stable/auto_examples/model_selection/plot_roc.html}{scikit-learn}.
\end{Rem}

\begin{paragraph}{Discussion.} Choosing the right evaluation measure in multi-class classification is crucial for understanding how well a model performs across various classes. Average accuracy extends the traditional accuracy measure to multiple classes by averaging performance across each class. While this provides a simple overview, it does not account for class imbalances, making it less reliable when class sizes differ significantly.

The micro F$_\beta$-score aggregates the true positives, false positives, and false negatives across all classes to compute an overall evaluation measure. While this approach provides a single summary statistic of the model's performance, it tends to favor dominant classes in imbalanced datasets. It treats each instance equally, regardless of its class, potentially obscuring the performance in less frequent or minority classes. As a result, micro F$_\beta$-score may not be the best choice for evaluating models in scenarios where the performance of minority classes is critical.

\end{paragraph}

\subsection{Measures on Imbalanced Data with Single Class}\label{section_classification_imbalanced_single_class}

In binary classification, imbalanced data occurs when one class significantly outnumbers the others, leading to biased model performance where the majority class dominates. Standard measures like accuracy can be misleading, as high accuracy might reflect the majority class prediction. Specialized measures, such as precision-recall curves and cost-sensitive methods, are crucial for evaluating model performance in these cases. Chance correction, which adjusts measures like the matthews correlation coefficient (MCC) to account for imbalance, helps provide a more accurate assessment of model effectiveness across both classes. The following sections cover fundamental and advanced measures for imbalanced single-class classification tasks.

\textbf{Balanced accuracy} is an evaluation measure used to evaluate binary classifiers, particularly in situations with imbalanced data \cite{bekkar2013evaluation}. Unlike standard accuracy from Sec.~\ref{sec_measures_balanced_single_class}, which can be biased towards the majority class, balanced accuracy adjusts for class imbalance by averaging the recall and precision across classes. The balanced accuracy $A_\text{balanced}$ is defined as the arithmetic mean of recall (TPR) and the true negative rate $TNR$.
\begin{align}\label{balanced_accuracy}
	A_\text{balanced} = \frac{TPR + TNR}{2} = \frac{1}{2}\,\left(\frac{TP}{FN + TP} + \frac{TN}{TN + FP}\right).
\end{align}

The balanced accuracy is particularly useful when the dataset is imbalanced, as it ensures that both classes contribute equally to the evaluation measure, regardless of their frequencies. This makes it a robust measure for evaluating classifiers in such contexts. However, if the classes are extremely imbalanced towards one class, the balanced accuracy is unsuitable as an evaluation measure overestimating the minority class in the mean. 

In comparison, the \textbf{matthew correlation coefficient (MCC)},  also known as \textbf{phi coefficient}, incorporates all four entries of the confusion matrix and conducts a reliable evaluation measure for imbalanced data \cite{Chicco2020}. It was introduced as \cite{MCC}: 
\begin{equation}\label{MCC}
MCC = \frac{TP*TN - FP*FN}{\sqrt{(TP+FP)*(TP+FN)*(TN+FP)*(TN+FN)}}
\end{equation}
and can be thought of as a coefficient between the observed and predicted binary classifications with values between $-1$ and $1$. If the coefficient reaches a value of $1$, it is an indicator for perfect classification, and $-1$ represents a total disagreement with the predictions. Therefore, a random classifier would have a correlation coefficient of $MCC = 0$ since the nominator (product of all correct predictions minus the product of all wrong predictions) will be zero \cite{chicco2020advantages}. Therefore, MCC is a chance-corrected measure.

Likewise, \textbf{cohen's kappa} is a chance-corrected measure used to evaluate classification performance, particularly useful in scenarios where class imbalances may affect standard accuracy measures \cite{GreveWentura_1997}. It adjusts for the agreement that could occur by chance, providing a more robust measure of classifier performance. It is defined as 
\begin{align}
    \kappa = \frac{p_0 - p_e}{1 - p_e},
\end{align}
where \( p_0 \) represents the probability of agreement or correct classification, determined as the sum of the diagonal elements in the confusion matrix divided by the total number of events (i.e., the accuracy). Further, \( p_e \) represents the hypothetical probability of chance agreement, corresponding to the case where the model classifies randomly. The value of \( p_e \) is calculated as \( p_e = p_{\text{correct}} + p_{\text{incorrect}} \) with
\begin{equation}\label{cohens_kappa}
\begin{split}
	p_{\text{correct}} &= \frac{(TP + FP)(TP + FN)}{(TP + FP + FN + TN)^2}, \\
	p_{\text{incorrect}} &= \frac{(FN + TN)(FP + TN)}{(TP + FP + FN + TN)^2}.
\end{split}
\end{equation}
Cohen's kappa is a more generally applicable evaluation measure than, e.g., the accuracy due to the handling of imbalanced data. Unlike the balanced accuracy, the chance correction in cohen's kappa accounts for the agreement expected by random chance, similar to the MCC. However, the MCC is more sensitive to class imbalance than cohen's kappa in certain situations, making it potentially more informative~\cite{chicco2021thematthews}.

Like cohen's kappa, \textbf{scott's pi}~\cite{japkowicz2011evaluating} is a chance-corrected measure to evaluate the agreement between two raters or classifiers based on the observed class distributions and is defined as
\begin{equation}\label{scotts_pi}
	\pi = \frac{p_0 - p^S_e}{1 - p^S_e},
\end{equation}
where $p^S_e$ is the sum of squared joint proportions, given by
\begin{equation}
    p^S_e = \left(\frac{TP + FN + TP + FP}{2*(TP + TN + FN + FP)}\right)^2 + \left(\frac{FP + TN + FN + TN}{2*(TP + TN + FN + FP)}\right)^2.
\end{equation}

\begin{wrapfigure}[15]{r}{0.4\textwidth}
  \begin{center}
  \vspace{-0.7cm}
    \includegraphics[width=0.3\textwidth]{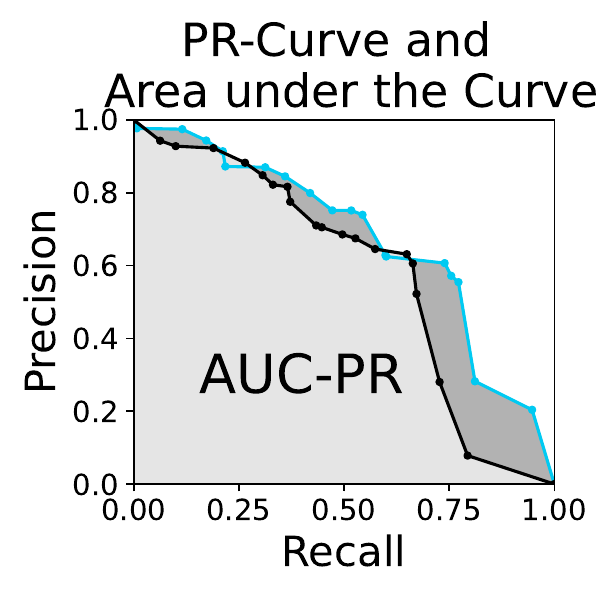}
  \end{center}
  \caption{The area under the precision-recall curve (AUC-PR) for two models illustrates the correlation between the precision and recall for a model w.r.t.~different decision thresholds.
  \label{fig:PR_example}}
\end{wrapfigure}

A common graphical evaluation measure is the \textbf{area under the precision-recall curve (AUC-PR)} \cite{saito2015precision}. It is given by the trade-off between precision and recall for various decision thresholds. Compared to the AUC-ROC from Sec.~\ref{sec_measures_balanced_single_class}, it is more sensitive to performance on imbalanced datasets as it specifically evaluates the precision and recall for the positive class. In contrast, the AUC-ROC considers both true positive and false positive rates, which can lead to overly optimistic assessments when the data is highly imbalanced. The higher the AUC-PR score, the better the classifier, cf.~Fig.~\ref{fig:PR_example}, where the red curve shows the better model, thus having a higher AUC-PR.

Unlike AUC-ROC, which treats all classification errors equally, the \textbf{H-measure} considers the specific costs of the different misclassification errors \cite{hand2023notes}. This makes it especially useful in situations where the impact of misclassifications is not uniform. 
The H-measure is determined as the expectation of the misclassification cost $L(\tau)$ for given thresholds $\tau$, based on the costs of false positives $c(\text{FP})$ and false negatives $C(\text{FN})$:
\begin{align}\label{h_measure}
H &= \int\limits_{0}^{1} L(\tau) \, p(\tau) \, d\tau \quad\text{ with }\quad \\
L(\tau) &= p \cdot c(\text{FP}) \cdot F_0(\tau) + (1 - p) \cdot c(\text{FN}) \cdot (1 - F_1(\tau)), \nonumber
\end{align}
where $p$ is the prior probability of the positive class, and $F_0(\tau)$ and $F_1(\tau)$ are the cumulative distribution functions of scores for the negative and positive classes. Thereby, the H-measure can account for the varying costs of different types of misclassification errors, providing a more accurate and context-sensitive evaluation of classifier performance.

The \textbf{likelihood ratio}, in turn, differs from the H-measure by focusing on the ratio of probabilities of making correct versus incorrect decisions rather than adjusting for misclassification costs. 
It is expressed as the positive likelihood ratio $LR_+$ for the positive class and the negative likelihood ratio $LR_-$ for the negative class. The positive likelihood ratio is defined as $LR_+ = \frac{TPR}{1 - TNR}$ representing the probability of a true positive (TP) event relative to the probability of a false positive (FP) event. Conversely, the negative likelihood ratio is defined as $LR_- = \frac{1 - TPR}{TNR}$,
which gives the ratio of the probability of a false negative (FN) event to the probability of a true negative (TN) event and thus does not allow for chance correction.

A higher $LR_+$ indicates a better model performance, as it suggests a higher probability of correct positive predictions, while a lower $LR_-$ indicates better performance by minimizing incorrect negative predictions. The concept and application of likelihood ratios are well-established in evaluating diagnostic tests and are extensively discussed in \cite{fletcher1982clinical}.

\begin{paragraph}{Discussion.}
    When dealing with imbalanced data in single-class classification, selecting the proper evaluation measures is crucial, as standard measures like accuracy often provide a misleading picture of model performance. Balanced accuracy offers a straightforward adjustment to traditional accuracy by accounting for class imbalance. It ensures that both classes contribute equally to the evaluation measure, making it useful in moderately imbalanced scenarios. However, it may still be insufficient in cases of extreme imbalance, as it can under-represent the minority class.
    
    MCC provides a more holistic view by explicitly considering all elements of the confusion matrix, making it particularly robust in situations with severe class imbalances. Its ability to capture the relationship between true positives, true negatives, false positives, and false negatives makes it more reliable than balanced accuracy in such contexts. Additionally, MCC is inherently chance-corrected, as its formula adjusts for predictions that could occur randomly, further enhancing its robustness in imbalanced datasets.

    Cohen's kappa also introduces chance correction, offering a finer classifier performance evaluation by accounting for the agreement expected by random chance. Applying chance correction makes cohen's kappa valuable when comparing models on imbalanced data, though it may be less sensitive to extreme imbalances than MCC. Scott's pi, similar to cohen's kappa, adjusts for chance but is tailored to the observed class distribution, providing another perspective on model performance. However, like cohen's kappa, its usefulness diminishes in cases of extreme imbalance, where more sensitive measures like MCC may be preferred.
    
    Precision-recall curves and the AUC-PR measure are particularly effective for imbalanced data, as they focus on the performance of the positive class. Determining the minority class as the positive class, the PR-curve and AUC-PR allow for a higher penalty of false classifications in the minority class.
    
    The likelihood ratio provides a different approach by focusing on the ratio of probabilities of making correct versus incorrect decisions rather than adjusting for misclassification costs like the H-measure. The positive likelihood ratio ($LR_+$) assesses the probability of a true positive relative to a false positive. In contrast, the negative likelihood ratio ($LR_-$) evaluates the probability of a false negative relative to a true negative. A higher $LR_+$ indicates better performance for positive predictions, and a lower $LR_-$ signifies better performance for negative predictions. This measure is particularly valuable in diagnostic testing and scenarios where understanding the odds of correct versus incorrect classifications is essential.
    
    Finally, the H-measure provides a sophisticated approach by incorporating the costs associated with different misclassification errors. Therefore, applications where the consequences of false positives and false negatives are unequal profit from the H-measure. However, its complexity makes it less accessible for general use when misclassification costs are not well-defined or vary significantly.
    
    In summary, the choice of measure should be guided by the specific characteristics of the dataset and the importance of different types of errors. MCC and AUC-PR are often the most reliable for general-purpose evaluation in imbalanced datasets. In contrast, measures like balanced accuracy and cohen's kappa may be better suited for less extreme imbalances or when a simple chance correction is needed. The H-measure, while powerful, is most appropriate in contexts where the costs of errors are clearly understood and can be accurately quantified. The likelihood ratio is beneficial when the goal is to understand the probabilities of making correct versus incorrect decisions, adding another layer of insight to model evaluation, especially in diagnostic applications.
    
\end{paragraph}

\subsection{Measures on Imbalanced Data with Multi-Class}\label{section_imbalanced_multiclass}

Imbalances in multi-class data are particularly challenging and require sophisticated evaluation measures. One approach is to enhance the previously defined micro measures to \textbf{macro precision} \( P_\text{macro} \) and \textbf{macro recall} \( R_\text{macro} \) by averaging the per-class scores \cite{sokolova2009systematic}. These metrics provide an equal-weighted view of class-wise performance, regardless of class imbalance:
\begin{align}\label{makro_precision}
    P_\text{macro} &= \frac{1}{C} \sum_{i=1}^C \frac{TP_i}{TP_i + FP_i}, \quad
    R_\text{macro} = \frac{1}{C} \sum_{i=1}^C \frac{TP_i}{TP_i + FN_i}.
\end{align}

In other words, macro precision and recall are computed by first assessing the evaluation measures for each class independently and then averaging these scores over all classes. Considering the individual class scores makes the measures particularly useful when classes are of varying sizes or importance, as they prevent dominant classes from overshadowing the evaluation.

The \textbf{macro F$_\beta$-score}, which generalizes the F$_\beta$-score to allow for adjustable weighting between precision and recall, is analogously defined as:
\begin{align}\label{macro_f_beta}
    F_{\beta,\text{macro}} = \frac{(1+\beta^2) P_\text{macro} R_\text{macro}}{\beta^2 P_\text{macro} + R_\text{macro}}.
\end{align}

Another widely used metric for multi-class evaluation is the \textbf{balanced accuracy}, which can be extended to multi-class scenarios as:
\begin{align}\label{balanced_accuracy_macro}
    A_\text{balanced} &= \frac{TPR_\text{macro} + TNR_\text{macro}}{2} \\  &= \frac{1}{2C}\,\sum_{i=1}^C\left(\frac{TP_i}{FN_i + TP_i} + \frac{TN_i}{TN_i + FP_i}\right), \nonumber
\end{align}
where we utilize the macro true positive rate (macro TPR) and the macro true negative rate (macro TNR).
Balanced accuracy is particularly suited for imbalanced datasets, as it combines both the recall (TPR) and the true negative rate (TNR), providing a robust measure of performance in scenarios with class imbalance.

For a graphical evaluation measure in highly imbalanced multi-class problems, the \textbf{AUC-PR} (area under the curve - precision-recall) is recommended. By utilizing \( P_\text{macro} \) and \( R_\text{macro} \) instead of \( P \) and \( R \), it effectively summarizes the trade-off between precision and recall across different thresholds, accounting for the varying importance of classes. Therefore, the AUC-PR is a practical choice when negatives significantly outnumber the positive instances.

The above measures for imbalanced multi-class problems do not allow chance correction. However, chance-corrected measures model explicitly the performance achievable by chance, thus allowing for expressing the proportion of model performance against the achievement of random guessing. Therefore, some measures of single-class classification have been extended to utilize chance correction for multiple classes. 

The \textbf{cohen's kappa$_n$} \cite{Hsu01082003} is the general case of the S-coefficient and cohen's kappa from Subsec.~\ref{section_classification_imbalanced_single_class}, expressing the proportion of expert knowledge to the performance of chance, defined as
\begin{align}\label{cohens_kappa_n}
    \kappa_n =\frac{p_0-p_e}{1-p_e},
\end{align}
with the base rate agreement $p_e=\frac{1}{C}$ for $C$ different classes. It directly applies to multiple imbalanced classes by considering the number of classes in the random performance.

Besides, the \textbf{cohen's weighted kappa} \cite{ben2008comparison} uses weights per rater agreement pair, i.e., per entry of the confusion matrix $c_{i,j}$ the agreement probability $p_{i,j} = \frac{c_{i,j}}{|C|}$ is weighted in the cohen's kappa with predefined weights $w_{i,j}$ (where $w_{i,j}=w_{j,i}$):
\begin{align}\label{cohens_kappa_weighted}
    \kappa_w = \frac{\displaystyle \sum_{i,j}w_{i,j}p_{i,j} - \sum_{i,j} w_{i,j} p_i p_j}{1-\displaystyle \sum_{i,j} w_{i,j} p_i p_j},
\end{align}
using the marginals $p_i=\frac{1}{|C|}\sum_j c_{i,j},\ p_j=\frac{1}{|C|}\sum_i c_{i,j}$. The weights here enable considering cost-sensitive classification problems for multiple imbalanced classes.

The matthews correlation coefficient (MCC) has been extended to the multi-class case in the form of the $\bm{R_C}$ statistics for C different classes, providing a comprehensive measure of classification performance in multi-class scenarios. This extension considers the interactions between all classes, capturing both the correct classifications and the misclassifications across the entire confusion matrix. The $\bm{R_C}$ statistics are defined as follows:
\begin{align}\label{R_C statistics}
    R_C = \frac{\displaystyle \sum_{i,j,k} c_{i,i}\cdot c_{j,k}- c_{i,j}\cdot c_{k,i}}
    {\displaystyle \sqrt{\sum_i\sum_j c_{i,j} \left(\sum_{i'\mid i'\neq i} \sum_{j'} c_{i', j'}\right)}\cdot \sqrt{\sum_i\sum_j c_{j,i} \left(\sum_{i'\mid i'\neq i} \sum_{j'} c_{j', i'}\right)}}.
\end{align}
Here, $\sum_i c_{i,i}$ is the total number of correctly classified samples, $\sum_{j,k} c_{j,k}$ is the total number of samples, $\sum_{i} c_{i,j}$ the number of times class $j$ truly occurred, and $\sum_{j} c_{i,j}$ the number of times class $i$ has been predicted.
The $R_C$ statistics is bounded by the interval $[-1,1]$, with higher values indicating better classification performance.

\paragraph{Discussion.} 
Evaluating measures for imbalanced multi-class datasets necessitates careful attention to class distributions.
Evaluation measures such as macro precision and macro recall average scores across all classes offer a balanced perspective suitable for datasets with varying class sizes or importance. 
However, these evaluation measures may diminish the prominence of dominant classes, potentially underestimating their influence in specific contexts. 

Balanced accuracy, which combines recall and the true negative rate (TNR), effectively addresses this issue by giving equal consideration to minority classes, making it a practical choice in datasets with substantial class imbalances.

Graphical evaluation measures like AUC-PR, adapted for multi-class scenarios, clearly represent the trade-offs between precision and recall, particularly highlighting performance on positive instances in highly imbalanced datasets. 

Chance-corrected metrics, such as the generalized cohen's kappa and the matthews correlation coefficient (RC statistic), account for random agreement, offering a more precise evaluation of model performance. While these measures are more computationally intensive, they are especially valuable in settings where the impact of random classification needs to be minimized.

\begin{center}
\begin{tikzpicture}[
    grow cyclic,
    every node/.style = {align=center},
    level distance=2.7cm, 
    sibling distance=6cm 
]

\node[root] {\hyperref[section_classification]{\Large{\underline{\textbf{Classification}}}}}
    child[grow=up] {node[decisionBIG] {\large \textbf{Balanced}}
        child[grow=135] {node[decisionMittel] {\hyperref[sec_measures_balanced_single_class]{\large Single-Class}}
            child[grow=180] {node[decision] {\normalsize Type-1 \\ Appropriate}
                child[grow=225] {node[leaf] {\hyperref[precision]{\scriptsize Precision}}}
                child[grow=180] {node[leaf] {\hyperref[precision]{\scriptsize FDR}}}
                child[grow=135] {node[leaf] {\hyperref[TNR]{\scriptsize TNR}}}
            }
            child[grow=135] {node[decision] {\normalsize Type-2 \\ Appropriate}
                child[grow=135] {node[leaf] {\hyperref[NPV]{\scriptsize Recall}}}
                child[grow=90] {node[leaf] {\hyperref[NPV]{\scriptsize NPV}}}
            }
            child[grow=90] {node[decision] {\normalsize Type-1\& \\Type-2}
                child[grow=90] {node[leaf] {\hyperref[f_beta]{\scriptsize $F_\beta$ Score}}}
            }
            child[grow=225] {node[decision] {\normalsize Graphical}
                child[grow=225] {node[leaf] {\hyperref[fig:ROC_example]{\scriptsize AUC-ROC}}}
                child[grow=270] {node[leaf] {\hyperref[fig:DET_example]{\scriptsize DET Curve}}}
            }
            child[grow=45] {node[leaf] {\hyperref[accuracy]{\scriptsize Accuracy}}}
            child[grow=270] {node[leaf] {\hyperref[accuracy]{\scriptsize Error Rate}}}
        }
        child[grow=45] {node[decisionMittel] {\hyperref[sec_measures_balanced_multi_class]{\large Multi-Class}}
        child[grow=90] {node[decision] {\normalsize Type-1 \\ Appropriate}
        child[grow=45] {node[leaf] {\hyperref[micro_precision_micro_recall]{\scriptsize Micro Precision}}
        }}
        child[grow=45] {node[decision] {\normalsize Type-2 \\ Appropriate}
        child[grow=0] {node[leaf] {\hyperref[micro_precision_micro_recall]{\scriptsize Micro Recall}}}
        }
        child[grow=0] {node[decision] {\normalsize Type-1 \& \\ Type-2}
        child[grow=0] {node[leaf] {\hyperref[micro_f_beta]{\scriptsize Micro F$_\beta$ Score}}}
        }
        child[grow=310] {node[decision] {\normalsize Graphical}
        child[grow=315] {node[leaf] {\hyperref[rem_graphical_extension]{\scriptsize AUC-ROC}}}
        child[grow=0] {node[leaf] {\hyperref[rem_graphical_extension]{\scriptsize DET Curve}}}
        }
        child[grow=270] {node[leaf] {\scriptsize \hyperref[average_accuracy]{Average Accuracy}}} 
        }}
    child[grow=down] {node[decisionBIG] {\large \textbf{Imbalanced}}
        child[grow=225] {node[decisionMittel] {\hyperref[section_classification_imbalanced_single_class]{\large Single-Class}}
            child[grow=225] {node[decision] {\normalsize No Chance Correction}
                child[grow=225] {node[leaf] {\hyperref[balanced_accuracy]{\scriptsize Balanced Accuracy}}}
                child[grow=180] {node[leaf] {\hyperref[h_measure]{\scriptsize Likelihood ratio}}}
            }
            child[grow=135] {node[decision] {\normalsize Chance-Corrected}
                child[grow=180] {node[leaf] {\hyperref[cohens_kappa]{\scriptsize Cohen's Kappa}}}
                child[grow=225] {node[leaf] {\hyperref[scotts_pi]{\scriptsize Scott's Pi}}}
                child[grow=45] {node[leaf] {\hyperref[h_measure]{\scriptsize H measure}}}
                child[grow=135] {node[leaf] {\hyperref[MCC]{\scriptsize MCC}}}
            }
            child[grow=270] {node[decision] {\normalsize Graphical}
                child[grow=270] {node[leaf] {\hyperref[fig:PR_example]{\scriptsize AUC-PR}}}
            }
        }
        child[grow=315] {node[decisionMittel] 
        {\hyperref[section_imbalanced_multiclass]{\large Multi-Class}}
            child[grow=45] {node[decision] {\normalsize Chance-Corrected}
                child[grow=45] {node[leaf] {\hyperref[cohens_kappa_weighted]{\scriptsize $\kappa_w$}}}
                child[grow=0] {node[leaf] {\hyperref[R_C statistics]{\scriptsize $R_C$ statistics}}}
                child[grow=135] {node[leaf] {\hyperref[cohens_kappa]{\scriptsize$\kappa_n$}}}
            }
            child[grow=315] {node[decision] {\normalsize No Chance-Corrected}
                child[grow=0] {node[leaf] {\hyperref[macro_f_beta]{\scriptsize Macro F$_\beta$ Score}}}
                child[grow=45] {node[leaf] {\hyperref[makro_precision]{\scriptsize Macro Recall}}}
                child[grow=270] {node[leaf] {\hyperref[makro_precision]{\scriptsize Macro Precision}}}
                child[grow=315] {node[leaf] {\hyperref[balanced_accuracy_macro]{\scriptsize Balanced Accuracy}}}
            }
            child[grow=270] {node[decision] {\normalsize Graphical}
                child[grow=270] {node[leaf] {\hyperref[balanced_accuracy_macro]{\scriptsize AUC-PR}}}
            }
        }
};

\end{tikzpicture}
\end{center}

\newpage
\section{Clustering Measures}\label{section_clustering}

Clustering evaluation differs from supervised learning metrics as it analyzes whether identified clusters fit the data well, either based on given labels or the structure of the clustering. Different absolute clustering measures have been developed in the literature. At the end of this section, we provide an overview-giving tree of the clustering metrics.

The \textbf{cluster purity} \(C_\text{purity}\) is used when ground truth labels are available. It measures the proportion of correctly classified instances in each cluster:
\begin{align}\label{C_purity}
    C_\text{purity} = \frac{1}{n} \sum_{i = 1}^k \max_{j \in \{1,...,k\}} |C(x_i) \cap \bar{C}_j|,
\end{align}
where \(C(x_i)\) denotes the predicted cluster of sample $x_i$, \(\bar{C}_j\) represents the $j$-th ground truth cluster, $k$ the number of clusters and $n$ is the number of samples in total. 
This metric is only valid when comparing models with the same number of clusters, as higher cluster purity is naturally obtained with more clusters, e.g., the model classifies each sample in its cluster, and $C_\text{purity}$ will reach $1$\,\cite{Rao2018exploring}.

Similarly, the \textbf{rand index} \(R\) evaluates the similarity between ground truth and predicted clusters by considering similar assignments of data pairs to either the same cluster or different clusters. Considering a data set $\mathcal{D}=\{x_1,\ldots,x_n\}$ with $n$ elements, $k$ ground truth clusters $\bar{C} = \{\bar{C}_1, ..., \bar{C}_k\}$ and the predicted clusters $\mathcal{C}=\{C_1,\ldots, C_l\}$, then the rand index is defined as
\begin{align}\label{R}
R = \frac{a + b}{\binom{n}{2}},
\end{align}
where \(a\) and \(b\) represent the counts of agreements between the ground truth and predicted clusters w.r.t. assigning a data pair to a common cluster or to different clusters \cite{rand1971objective}. More precisely, $a$ is the number of point pairs assigned to the same cluster in the prediction and the ground truth
\begin{align}
    a=|\{(x_i,x_j)\in\mathcal{D}\mid \exists m\in[k],m'\in[l]\,:\, x_i,\,x_j\in C_m \wedge x_i,\,x_j\in \bar{C}_{m'}\}|,
\end{align}
and $b$ is the number of point pairs assigned to different clusters in the prediction and ground truth 
\begin{align}
    b=|\{(x_i,x_j)\in\mathcal{D}\mid \exists m\in[k], m'\in[l]\,:\, \left(x_i\in C_m \wedge \,x_j\notin C_m \right) \wedge \left(x_i\in \bar{C}_{m'} \wedge \,x_j\notin \bar{C}_{m'} \right)\}|.
\end{align}

An ideal clustering model where the predicted clusters coincide with the ground truth scores \(R = 1\), while $R=0$ if one clustering (the groud truth or the prediction) assigns all data points to one cluster and the other clustering consists of one cluster per point \cite{wagner2007comparing}. In general, 
$a + b$ can be interpreted as the number of agreements between ground truth clusters and predicted clusters, i.e., as the $TP$ and $TN$ entries of a binary confusion matrix diagonal. Therefore, the rand index can be interpreted as the accuracy of a binary classification and consequently can be overly optimistic and often nearing $1$ even with suboptimal clustering.

To address this issue, the \textbf{adjusted rand index} (ARI) refines the rand index to correct for chance agreement by integrating the expected rand index \(\mathbb{E}(R)\) \cite{hubert1985comparing}:
\begin{align}\label{ARI}
\text{ARI} = \frac{R - \mathbb{E}(R)}{\text{max}(R) - \mathbb{E}(R)}, 
\end{align}
where $\text{max}(R)=1$. The ARI ranges from \(-1\) to \(1\) where $ARI=1$ is reached with a perfect agreement, $ARI=-1$ is total disagreement with expected rand index of $0.5$, and \(0\) corresponds to a random clustering. Unlike the rand index, ARI provides a more accurate measure by accounting for random clustering similarities, making it particularly useful in evaluating clustering quality when the number of clusters or class balance varies.

Furthermore, the \textbf{fowlkes-mallow index} \cite{fowlkes1983method,meilua2007comparing} scores a clustering based on the entries of the confusion matrix. Given the ground truth, the fowlkes-mallow index is then defined as
\begin{align}\label{FM}
    FM = \sqrt{\frac{TP}{TP+FP}\cdot\frac{TP}{TP+FN}}
\end{align}
considering the recall and precision of the clustering performance. For perfectly clustered data, the $FM$ index is $1$; for completely unrelated clusterings, the index is $0$.

For unsupervised clustering where the ground truth is unavailable, the \textbf{silhouette coefficient} provides insight into the clustering quality based on the data structure. It evaluates how well-separated and cohesive the clusters are by comparing the average distance between points within the same cluster (a) and the nearest cluster (b):
\begin{align}\label{s}
    s = \frac{b - a}{\max(a, b)}.
\end{align}

The coefficient ranges from -1 to 1, with values closer to 1 indicating well-separated, dense clusters, values around 0 suggesting overlapping clusters, and values close to -1 indicating poorly defined clusters. Aggregating silhouette scores across samples offers a comprehensive view of the clustering configuration.

\paragraph{Discussion.} 
When evaluating clustering models, the choice of measure depends heavily on the availability of ground truth labels and the specific objectives of the analysis. 

Cluster purity is straightforward and intuitive when ground truth labels are available, as it directly measures the proportion of correctly assigned instances within each cluster. However, it has a significant limitation: it tends to increase with the number of clusters, often leading to overly optimistic evaluations in models that generate many small clusters. Therefore, cluster purity is most useful when models produce the same number of clusters, allowing for a fair comparison.

The rand index (RI) provides a more nuanced evaluation by considering all pairs of points and assessing whether they are consistently clustered in predicted and ground truth clusters. While RI helps understand the overall agreement between the clustering result and the ground truth, it tends to yield high values even when the clustering is suboptimal, especially in imbalanced datasets.

The adjusted rand index (ARI) addresses the optimism inherent in RI by adjusting for the similarity expected by chance. Therefore, the ARI is more reliable when the number of clusters varies, or the class distribution is unbalanced. The ability to penalize chance agreements in the ARI makes it preferable over RI when the primary concern is to assess clustering quality without the bias introduced by random assignments.

The fowlkes-mallow index is a symmetric measure to evaluate the agreement between two clusterings. In the supervised clustering context, the index indicates the coincidence of the clustered data assignment with the ground truth. It is used when the pairwise relationship of two clusterings is considered and independent of the number of clusters. However, when the clusters are imbalanced, the FM index disregards different cluster sizes and thus may be high even if small clusters are incorrectly assigned. 

In contrast, the silhouette coefficient is tailored for unsupervised clustering scenarios with no ground truth labels. It provides insights into the cohesion and separation of clusters, with higher scores indicating well-defined clusters. The silhouette coefficient is particularly useful in exploratory data analysis, which aims to identify natural groupings within the data. However, it may be less informative in datasets with overlapping clusters or varying densities, where the metric could yield ambiguous results.

\begin{center}
\begin{tikzpicture}[
    grow cyclic,
    every node/.style = {align=center},
    level distance=2.3cm, 
    sibling distance=7cm 
]

\node[root] {\hyperref[section_clustering]{\large \underline{\textbf{Clustering}}}}
    child[grow=225] {node[decisionBIG] {\normalsize Ground Truth Available}
        child[grow=225] {node[leaf] {\hyperref[C_purity]{\scriptsize Cluster Purity}}}
        child[grow=270] {node[leaf] {\hyperref[ARI]{\scriptsize (Adjusted) Rand Index}}} 
        child[grow=315] {node[leaf] {\hyperref[FM]{\scriptsize Fowlkes-Mallow Index}}} 
    }
    child[grow=315] {node[decisionBIG] {\normalsize Ground Truth Unknown}
        child[grow=315] {node[leaf] {\hyperref[s]{\scriptsize Silhouette Coefficient}}}
    };
\end{tikzpicture}
\end{center}

\section{Ranking Measures}\label{section_ranking}
Ranking measures are used for recommendation systems~\cite{ricci2022recommender,liu2018STAMP}, information retrieval (e.g., search engines) \cite{liu2009learning}, drug discovery \cite{agarwal2010ranking}, and other fields where the objective is to identify the top-N items in a dataset. The evaluation measures used in these contexts can vary based on the emphasis on different types of errors and are gathered in a concluding concept tree providing a clear overview.

\textbf{Precision@k} (or \textbf{hits@k}) is used to minimize false positives, making it suitable for recommendation systems and search queries where irrelevant items reduce user satisfaction. Precision@k, denoted as $P@k$, is defined as the fraction of relevant items among the top-$k$ recommendations:
\begin{equation}\label{precision_at_k}
P@k = \frac{\#\,\text{relevant items in top-} k}{k}.
\end{equation}
However, the performance measured by $P@k$ can decline with larger values of $k$, especially if the total number of relevant items is low. Additionally, $P@k$ treats all positions equally, which may not align with user behavior where higher-ranked items are more significant~\cite{Craswell2009}.

To address the dependency on $k$, \textbf{average precision@k} ($AP@k$) averages $P@k$ over all possible $k$ values, providing a more stable measure:
\begin{equation}\label{average_precision_at_k}
AP@k = \frac{1}{K} \sum_{i=1}^K P@k_i,
\end{equation}
where $K$ is the total number of possible ranks.

\textbf{Recall@k} is employed to minimize false negatives and is defined as the proportion of relevant items identified within the top-$k$ recommendations:
\begin{equation}\label{recall_at_k}
R@k = \frac{\#\,\text{relevant items in top-} k}{\#\,\text{total relevant items}}.
\end{equation}
Selecting an appropriate value of $k$ is critical, as $R@k$ can trivially reach 100\% if $k$ equals the total number of items in the dataset.

If the interest lies in the first relevant item, the \textbf{mean reciprocal rank (MRR)} is a useful measure. MRR calculates the average of the reciprocal ranks of the first relevant item across all samples $\mathcal{D}$:
\begin{equation}\label{MRR_clustering}
MRR = \frac{1}{|\mathcal{D}|} \sum_{x \in \mathcal{D}} \frac{1}{k_x},
\end{equation}
where $k_x$ is the rank of the first relevant item for sample $x$. MRR is particularly useful when only the top result matters, but it is less suitable when multiple relevant items are interesting.

For scenarios where ranking all relevant items is crucial, \textbf{(normalized) discounted cumulative gain (DCG)} is appropriate. DCG rewards models that rank relevant items higher, reflecting their importance in search queries and similar applications:
\begin{equation}\label{DCG_p}
DCG_p = \sum_{i=1}^p \frac{rel_i}{\log_2(i+1)},
\end{equation}
where $rel_i$ is the relevance of the item at rank $i$. The normalized DCG, nDCG, compares the DCG to the ideal DCG ($IDCG$):
\begin{equation}\label{nDCG_p}
nDCG_p = \frac{DCG_p}{IDCG_p},
\end{equation}
where $IDCG_p$ is computed by sorting all relevant items by their relevance up to position $p$. An nDCG value of 1.0 indicates a perfect ranking.

\paragraph{Discussion.}
When selecting ranking measures, it is crucial to consider the specific requirements of the task at hand, as each measure has advantages and disadvantages depending on the context.

Precision@k is a straightforward measure that minimizes false positives (type-1 error appropriate), making it highly suitable for applications where irrelevant items can significantly diminish user experience, such as recommendation systems and search queries. However, its limitation lies in treating all positions equally, which may not reflect actual user behavior, as users tend to prioritize items ranked higher. Additionally, precision@k can be misleading if the value of $k$ is not well-chosen, as it does not account for the total number of relevant items in the dataset.
Average precision@k (AP@k) improves upon precision@k by averaging the precision across all possible ranks, providing a more stable evaluation measure. Thereby, AP@k is particularly useful when the optimal value of $k$ is uncertain or when a single fixed $k$ might not adequately capture the model's performance. However, like precision@k, it assumes equal importance for all ranks, which may not always align with user preferences.

Recall@k, on the other hand, emphasizes minimizing false negatives (type-2 error appropriate), making it suitable for tasks where it is crucial to retrieve as many relevant items as possible, such as search engines or filtering systems. The primary drawback of recall@k is that it can easily reach its maximum value if $k$ is set too high, potentially leading to overestimating the model's performance. Therefore, a careful selection of $k$ is necessary to ensure a meaningful evaluation.

The mean reciprocal rank (MRR) is particularly valuable when the goal is to identify the first relevant item in a ranked list, such as in question-answering systems or scenarios where the top recommendation is of utmost importance. The MRR provides a transparent and interpretable measure that prioritizes finding the first correct answer. However, it may not be suitable when multiple relevant items are essential, as it only considers the rank of the first relevant item.

The normalized discounted cumulative gain (nDCG) is one of the most comprehensive ranking measures, as it accounts for the rank of all relevant items, giving more weight to items at higher ranks. Therefore, the nDCG is suitable for tasks like search engine optimization, where the user is more likely to interact with the top results. The primary advantage of nDCG is its ability to reflect the varying importance of ranks. Still, its complexity can be a drawback, particularly when compared to more straightforward measures like precision@k or MRR. Additionally, the nDCG requires a relevance score for each item, which may not always be easy to define.

\begin{center}
\begin{tikzpicture}[
    grow cyclic,
    every node/.style = {align=center},
    level distance=2.4cm, 
    sibling distance=4cm 
]

\node[root] {\hyperref[section_ranking]{\Large \underline{Ranking}}}
    child[grow=135] {node[decision] {\normalsize Type-1 Appropriate}
        child[grow=180] {node[leaf] {\hyperref[precision_at_k]{\scriptsize P@k}}}
        child[grow=225] {node[leaf] {\hyperref[average_precision_at_k]{\scriptsize AP@k}}}
    }
    child[grow=45] {node[decision] {\normalsize Type-2 Appropriate}
        child[grow=0] {node[leaf] {\hyperref[recall_at_k]{\scriptsize R@k}}}
    }
    child[grow=270] {node[decision] {\normalsize Emphasizing High Ranking}
        child[grow=0] {node[leaf] {\hyperref[MRR_clustering]{\scriptsize MRR}}}
        child[grow=180] {node[leaf] {\hyperref[nDCG_p]{\scriptsize nDCG}}}
        child[grow=135] {node[leaf] {\hyperref[DCG_p]{\scriptsize DCG}}}
    };

\end{tikzpicture}
\end{center}

\section{Domain-Specific Measures}\label{section_data-specific}

Based on the type of data, there are further evaluation measures that are tailored to domain-specific problems in particular. This section outlines key measures for assessing models across image, time series, text, and graph domains. 

\paragraph{Image-Specific Measures.}
Image datasets vary in importance depending on their application. For instance, while classifying animals in images might have low stakes, errors in medical image segmentation could lead to serious health consequences \cite{muller2022towards}. Although common Measures can be applied to image classification and segmentation tasks, image-specific measures like the \textbf{intersection-over-union (IoU)}, also known as the \textbf{jaccard/tanimoto similarity coefficient}, are often used in object detection and segmentation \cite{jaccard_tanimoto}. 

\paragraph{Text-Specific Measures.}
The evaluation of text-based models varies depending on the data encoding and the specific task. For natural language processing tasks such as text comprehension, standard measures like accuracy, precision, recall, and F1-score are often employed \cite[§3]{survey_reading_comprehension}. Additionally, the \textbf{exact match} metric assesses the percentage of system-generated text that matches the correct answer word for word. Advanced text-specific measures, including \textbf{ROUGE, BLEU, HEQ}, and \textbf{meteor}, combine or extend precision, recall, accuracy, and F1 to evaluate generated text more comprehensively.

\paragraph{Graph-Specific Measures.}
Graph learning tasks typically fall into node-level, edge-level, or (sub)graph-level categories, which can be further reduced to classical problems like classification or clustering. For example, node classification might use accuracy, while edge prediction might be evaluated using AUC or average precision. Dynamic node prediction could utilize measures like mean rank or mean reciprocal rank. Standard evaluation measures are commonly used to assess the resulting embeddings in graph neural networks, which often rely on encoder-decoder architectures. However, graph-specific tasks such as graph generation or reconstruction are evaluated based on the similarity between the original and generated or reconstructed graphs. Although various graph distance measures exist, they are rarely used for machine learning model evaluation and are generally not considered absolute measures \cite{graph_comarisons_metrics, evaluation_graph_generative_models}.

\section{Discussion and Conclusion}\label{section_discussion}

This paper broadly surveys absolute evaluation measures tailored to machine learning tasks like classification, clustering, and ranking. 
Our key contribution is to systematically categorize and discuss these evaluation measures based on the learning problem and the context of their applicability. Unlike prior surveys, we focus exclusively on absolute measures that offer practical evaluation of models independently of baselines and enhance reliability in cross-model comparisons.

Selecting suitable evaluation measures is crucial for effectively assessing machine learning models. Different measures yield distinct insights depending on problem type, data characteristics, and application objectives. We offer guidance on choosing appropriate measures by categorizing them according to their alignment with specific learning problems and considerations, such as chance correction and sensitivity to type-1 or type-2 errors, i.e., false positives and false negatives. Our survey proposes decision trees for metric selection, designed based on the challenges faced in each problem type.

In the case of classification, the class distribution and number of classes, as well as the priority of different errors, play a significant role in the choice of suitable evaluation measures. 
For single-class classification tasks on balanced data, accuracy provides a broad performance overview but becomes unreliable with increasing class imbalances. For balancing type-1 and type-2 errors,
the $\text{F}_\beta$-score is used, where $\text{F}_1$ weighs the two error types equally. For easy accessibility and a broader performance view, graphical tools like AUC-ROC are commonly used but risk over-optimism on imbalanced datasets. In the case of multi-class classification, average accuracy and micro $\text{F}_\beta$-scores were developed to provide a global evaluation measure across all classes.

Balanced accuracy partially adjusts for bias for imbalanced single-class data, while measures like matthews correlation coefficient (MCC) provide a more comprehensive perspective by considering all confusion matrix elements, making them more robust in extreme imbalance cases. Moreover, adjusted measures like cohen's kappa and scott's pi correct for chance agreement, reducing random impact on the performance evaluation. Precision-recall curves and AUC-PR focus on minority class performance, shifting the emphasis of misclassifications to the less represented class. Therefore, these graphical measures often outperform the AUC-ROC in imbalanced scenarios. Similar to this approach, the much more complex H-measure shifts the focus to the minority class by considering a misclassification cost. While these measures designed for imbalanced data can be applied to balanced scenarios, their actual importance emerges in contexts of class imbalance, where they offer nuanced insights that traditional metrics may overlook.

Clustering evaluation depends on the availability of ground truth. On the one hand, cluster purity and adjusted rand index (ARI) compare correctly classified instances per cluster using the ground truth, while ARI allows for a chance correction. On the other hand, the silhouette coefficient provides insights into cluster cohesion and separation for unsupervised clustering but may struggle with overlapping clusters or density variations.

In a ranking task, items are ordered concerning a selected objective. Ranking measures typically consider the first $k$ ranked items due to computational cost and simultaneously follow diverse goals. The precision@k minimizes the number of false positives. At the same time, recall@k focuses on maximizing the number of relevant items, and the mean reciprocal rank is high if many highly relevant items are among the first $k$ rankings. The normalized discounted cumulative gain evaluates rankings more comprehensively, balancing the positions and relevance of items in the ranking at the cost of increased complexity.

In a variety of specific contexts, domain-specific measures have been developed. For example, the intersection of union is commonly used for image segmentation, and it is particularly relevant for object detection applications. Text-based models are often evaluated with standard classification measures but may benefit from advanced measures like ROUGE and BLEU, which capture linguistic characteristics. Graph-based learning tasks comprise graph-, node, or edge-level tasks in classification or clustering. Therefore, graph-specific measures are based on classical evaluation measures, respectively.

Choosing the right evaluation measure for a specific learning task usually requires considering dataset characteristics, task objectives, and error priorities. Combining multiple measures yields a more holistic evaluation, aligning evaluation practices with model goals and data characteristics. 
This survey provides a structured approach to selecting appropriate evaluation measures, promoting clarity and reliability in assessing ML models across diverse applications.

\bibliographystyle{unsrt}
\bibliography{bibliography/bibliography}

\end{document}